# An Investigation on Support Vector Clustering for Big Data in Quantum Paradigm

Arit Kumar Bishwas[a, *], Ashish Mani[b], Vasile Palade[c]

[a] *AIIT, Amity University Uttar Pradesh, Noida, India, aritkumar.official@gmail.com*
[b] *EEE, ASET, Amity University Uttar Pradesh, Amity University, Noida, India, amani@amity.edu*
[c] *Faculty of Engineering, Environment and Computing, Coventry University, UK, vasile.palade@coventry.ac.uk*

**Abstract**

The support vector clustering algorithm is a well-known clustering algorithm based on support vector machines using Gaussian or polynomial kernels. The classical support vector clustering algorithm works well in general, but its performance degrades when applied on big data. In this paper, we have investigated the performance of support vector clustering algorithm implemented in a quantum paradigm for possible runtime improvements. We have developed and analyzed a quantum version of the support vector clustering algorithm. The proposed approach is based on the quantum support vector machine **[1]** and quantum kernels (i.e., Gaussian and polynomial). The classical support vector clustering algorithm converges in $O(M^2N)$ runtime complexity, where $M$ is the number of input objects and $N$ is the dimension of the feature space. Our proposed quantum version converges in $\sim O(logMN)$ runtime complexity. The clustering identification phase with adjacency matrix exhibits $O(\sqrt{M^3 lgM})$ runtime complexity in the quantum version, whereas the runtime complexity in the classical implementation is $O(M^2)$. The proposed quantum version of the SVM clustering method demonstrates a significant speed-up gain on the overall runtime complexity as compared to the classical counterpart.

## 1. INTRODUCTION

Clustering is a popular unsupervised machine-learning task, which groups input objects into multiple sets based on some similarities. There are many well-defined clustering algorithms, which work well on many practical problems. K-Means clustering **[2] [3]** is one of the most widespread clustering algorithms, but it has the drawback of needing to define the number of clusters in advance; although some upgraded versions, such as K-Means++ **[4]**, handle this limitation to some extent. Clustering approaches in **[5] [6]** show the advantages of using classification algorithms for clustering. One of the popular classical approaches is to use the one-class SVM (Support Vector Machines) and extend it to clustering problems, known as the support vector clustering method **[7]**. One of recent interesting progress in the field of quantum clustering discusses a distributed secure quantum machine learning protocol, which helps in classifying two-dimensional vectors to different clusters **[8]**. A higher number of dimensions will always be a tough problem to deal with in designing a clustering algorithm. Recently, the authors of **[9]** have discussed a new approach with the so-called quantum A-optimal projection (QAOP) algorithm. Dimensionality reduction is not discussed in this paper, however, it can be a promising technique and can be used with our proposed work. Our work focuses on the formulation of the quantum version of the support vector clustering (SVC) algorithm, but the above mentioned recent research works are interesting to address and explore in future work, which could be aligned with our current research investigations.

One-class SVM is an efficient way of estimating the density of a population **[10] [11]**. A transformation such that, $F: x \to F(x)$, helps in formulating the one-class SVM concept. Here, the function transforms the input object space into a higher dimensional feature space, such that the object points within dense localities are projected further from the origin of the assumed coordinate system. In the input object space, the support vectors outline the closed contours around the dense regions in the feature space, where a related decision function does the prediction as positive when the objects are inside the contour and negative elsewhere. This method is very useful in applications such as image retrieval, fault detection, context change detection, etc. **[12]**.

The support vector clustering method is based on one-class SVM and on using Gaussian/polynomial kernels. In support vector clustering, we form contours of input objects in a higher dimensional feature space. We replace the dot products of input feature vectors, say $\langle \vec{x}_i, \vec{x}_j \rangle$, in our one-class SVM formulation with a kernel function $K(\vec{x}_i, \vec{x}_j)$ – popularly known as the "kernel trick" - where $K$ may be a linear, polynomial or Gaussian kernel. These contour's boundaries are defined by support vectors and consist of a set of some specific input objects. We consider each contour boundary as a cluster. Once we define the contour boundaries, we can separate the clusters with the help of an adjacency matrix.

In this paper, we investigate the support vector clustering method in a quantum paradigm. In the proposed approach, we have used a one-class version of a quantum support vector machine with quantum kernels (i.e, quantum polynomial kernels as well as quantum Gaussian kernels **[13] [14]**) to design the support vector clustering algorithm. Our analysis shows that the proposed quantum version of support vector clustering shows significant performance gains (more than quadratic speed up gain in overall runtime complexity) as compared to the classical counterpart. This performance gain with a quantum version of support vector clustering (SVC) is significant especially when the input dataset is big data.

2. SUPPORT VECTOR MACHINES

2.1 *Classical Least Square SVM*

Support vector machines with the kernel trick is a very popular classification technique used for nonlinear datasets, where we first map the input objects into a higher dimensional feature space by using a kernel function. It then constructs an optimal separating hyperplane (with maximum separating margins) in the higher dimensional space to classify the data objects. In **[15],** the least square support vector machine (LS-SVM) has been discussed, which is based on the least square technique for function estimation **[16]**. In LS-SVM, instead of quadratic programming, we solve a linear system of equations in order to find the solutions. We formulate the problem by using equality constraints rather than inequality constraints.

With a given set of $M$ training objects $\{y_i, x_i\}_{i=1}^{M}$, where $x_i \in \mathbb{R}^N$ denotes the $i^{\text{th}}$ input and $y_i \in \mathbb{R}$ the $i^{\text{th}}$ output, the SVM objective is to construct a classifier of the form:

$$y(x) = sign\ \{\sum_{i=1}^{M} \alpha_i y_i K(x, x_i) + b\}. \tag{1}$$

where, $\alpha_i$ are support values, $b$ is a real constant and $K(x, x_i)$ is any kernel function. Let us suppose

$$w^T(x_i) + b \geq +1, \quad if\ y_i = +1 \tag{2}$$
$$w^T(x_i) + b \leq -1, \quad if\ y_i = -1, \tag{3}$$

where $\vec{w}$ is the normal vector to the hyperplane.

The above equations **(2 & 3)** can be written in the following equivalent formulation:

$$y_i[w^T(x_i) + b \geq 1], \quad \text{where } i = 1, \ldots, M \tag{4}$$

and the classifier is obtained as the solution to the following optimization problem:

$$\min_{w,b,e} \mathcal{J}_{LS}(w, b, e) = \frac{1}{2} w^T w + \gamma \frac{1}{2} \sum_{i=1}^{M} e_i^2 \tag{5}$$

which is subject to the following equality constraints:

$$y_i[w^T(x_i) + b] = 1 - e_i, \quad i = 1, \ldots, M \tag{6}$$

and $\gamma \frac{1}{2} \sum_{i=1}^{M} e_i^2$ is a penalty term, where $e_i$ is the error on example $i$ and $\gamma$ is the hyperparameter to tune the regularization versus the sum squared error. The Lagrangian function is then defined as:

$$\mathcal{L}(w, b, e; \alpha) = \mathcal{J}_{LS} - \sum_{i=1}^{M} \alpha_i [y_i \{w^T(\vec{x}_i) + b\} - 1 + e_i] \tag{7}$$

We can write the conditions for optimality as linear systems **[17-18]**, by taking partial derivatives of the Lagrangian function and eliminating the variables $e_i$ and $w$, and we get:

$$\begin{pmatrix} 0 & \vec{1}^T \\ \vec{1} & K + \gamma^{-1} \mathbb{I} \end{pmatrix} \begin{pmatrix} b \\ \vec{\alpha} \end{pmatrix} = \begin{pmatrix} 0 \\ \vec{y} \end{pmatrix} \tag{8}$$

$$\Rightarrow F \begin{pmatrix} b \\ \vec{\alpha} \end{pmatrix} = \begin{pmatrix} 0 \\ \vec{y} \end{pmatrix}. \tag{9}$$

where, $K_{ij} = K(\vec{x}_i^T, \vec{x}_j) = \vec{x}_i^T \cdot \vec{x}_j$ is the kernel matrix, $\mathbb{I}$ is the $M \times M$ identity matrix, $\vec{y} = (y_1, \ldots, y_M)^T$, $M$ components vector $\vec{1} = (1, \ldots, 1)^T$, $\vec{\alpha} = (\alpha_1, \ldots, \alpha_M)^T$. The support vector machine parameters are then determined by :

$$(b, \vec{\alpha}^T)^T = F^{-1}(0, \vec{y}^T)^T \tag{10}$$

Now, an unknown input object $\vec{x}$ can be classified by the following equation:

$$(0, \vec{y}^T)^T = F(b, \vec{\alpha}^T)^T \approx f(\vec{x}) = sgn\left(\sum_{j=1}^{M} \alpha_i \vec{y}_i K(\vec{x}_i, \vec{x}) + b\right) \tag{11}$$

We can extend the discussion to multiclass classification using two very popular approaches, i.e., "one-against-all" and "all-pair" **[19]** approaches. In the one-against-all approach, we first build and train $k$ quantum binary classifiers. Each of these quantum binary classifiers then classifies a given query state $|\vec{x}\rangle$ with some probability value. Then, the one-against-all algorithm finds the class for which the corresponding classifier's probability confidence score is highest, which is the predicted class.

In the all-pair approach, for each pair of classes, there is a binary classification problem, and hence we build $(k(k-1)/2)$ binary classifiers, where $k$ is the number of classes. Each binary classifier is then trained with associated training examples; therefore, we also define $(k(k-1)/2)$ sets of training examples. During prediction, we apply a voting mechanism, where we apply all $(k(k-1)/2)$ classifiers to an unseen data object and the class that got the highest number of " $+1$" predictions is the class predicted by the combined classifier.

## 2.2 Quantum Least Square SVM

The quantum version of least square support vector machines (binary or/and multiclass) has been discussed in **[13] [1] [20]**, which exhibits an exponential speed up as compared to the classical least square support vector machines (binary and multiclass). The quantum least squares support vector machine formulation allows us using the phase estimation and the quantum matrix inversion algorithm.

On the same lines as classical support vector machines, we formulate a general multiclass quantum SVM **[1]**. For the quantum formulation, we create the quantum states $|b_j, \vec{\alpha}_j\rangle$ by describing the hyperplane with the quantum matrix inversion algorithm. Therefore, the task is to solve the following equation **(12)** & **(13)**:

$$\hat{F}_j(|b_j, \vec{\alpha}_j\rangle) = |\vec{y}_j\rangle; j = 1, 2, 3, \dots k(k-1)/2 \qquad (12)$$

$$\Rightarrow |b_j, \vec{\alpha}_j\rangle = \hat{F}_j^{-1}(|\vec{y}_j\rangle), \qquad (13)$$

where $\hat{F}$ is the $(M+1) \times (M+1)$ dimensional normalized operator of $F$, and $M$ is the number of training examples. We now need to determine the quantum SVM parameters for the $j^{th}$ classifier, where $\hat{F}_j = \begin{pmatrix} 0 & \vec{1}^T \\ \vec{1} & \hat{K}_j + Y_j^{-1}\mathbb{I} \end{pmatrix}$, $\hat{K}_j$ is the kernel matrix for the $j^{th}$ classifier, $Y_j$ determines the relative weight of the SVM objective and the training error for $j^{th}$ classifier, $\vec{y}_j = (y_{j1}, \dots, y_{jM})^T$, $b_j$ are the biases, $\vec{\alpha}_j = (\alpha_{j1}, \dots, \alpha_{jM})^T$ are non-sparse vectors and act as the distance from the optimal margin for the $jth$ classifier, and $k$ is the number of classes. The classification of an unknown quantum state $|x\rangle$ is determined by the success probability $P_j^{(f,s)}$ (as shown in Table 1) of a swap test between $|b_j, \vec{\alpha}_j\rangle$ and $|\vec{x}\rangle$. $|\vec{x}\rangle$ will be classified as $+1$ or $-1$ with the quantum all-pair algorithm based on the following conditions **[1]**:

TABLE I. Probability conditions for classification

| Conditions | Classification of $|\vec{x}\rangle$ | Class classified as |
|---|---|---|
| $P_j^{(f,s)} < \frac{1}{2}$ | $+1$ | $f$ |
| $P_j^{(f,s)} \geq \frac{1}{2}$ | $-1$ | $s$ |

where $P_j^{(f,s)}$ characterizes the success probability of classifying data object as $f\ or\ s$ upon measurement by $j^{th}$ classifier.

The speedup gain is achievable during the training phase because of the quantum implementation of the matrix inversion algorithm **[21]**, non-sparse density matrices **[22]** and simulating sparse Hamiltonians **[23]**.

For solving, $\hat{F}_j(|b_j, \vec{\alpha}_j\rangle) = |\vec{y}_j\rangle; j = 1, 2, 3, \dots k(k-1)/2$, we determine the matrix exponential of $\hat{F}_j$. The $\hat{F}_j$ can be written as $\hat{F}_j = \frac{(J_j + K_j + \gamma_j^{-1}\mathbb{I}_j)}{trF_j}$, where $J_j = \begin{pmatrix} 0 & \vec{1}^T \\ \vec{1} & 0 \end{pmatrix}$ is a star graph, $K_j$ is the kernel matrix and $\gamma_j$ regulates the relative weight of training error and least square SVM objectives. We then obtain the following exponential:

$$e^{\frac{-i\hat{F}_j \Delta t}{trF}} = e^{\frac{-iJ_j \Delta t}{trF}} e^{\frac{-iK_j \Delta t}{trF}} e^{\frac{-i\gamma_j^{-1}\mathbb{I}_j \Delta t}{trF}} + O(\Delta t^2) \qquad (14)$$

where the eigenvectors and eigenvalues of the star graph $J_j$ are respectively $E_\pm^{J\_val} = \pm \frac{1}{\sqrt{2}}\left(|0\rangle \pm \frac{1}{\sqrt{M}} \sum_{r=1}^{M} |r\rangle\right)$ & $E_\pm^{J\_vec} = \pm \sqrt{M}$.

With several copies of the density matrix $\rho_j$, it is promising to implement $e^{-i\rho_j t}$ [24] for computing the matrix inverse $\widehat{K}_j^{-1}$, where $\widehat{K}_j$ is a non-sparse normalized Hermitian matrix. Based on the discussion in [24], the exponentiation runtime complexity is determined as $O(logN)$. $\widehat{K}_j$ is a normalized Hermitian matrix, so it is a potential candidate for quantum self-analysis [24]. We therefore evaluate $e^{-i\widehat{K}_j \Delta t}$ as

$$e^{-i\mathcal{L}_{\widehat{K}_j}\Delta t}(\rho) \approx \rho - i\Delta t[\widehat{K}_j, \rho] + O(\Delta t^2). \tag{15}$$

where, $\mathcal{L}_{\widehat{K}_j} = [\widehat{K}_j, \rho]$ and $N$ is the dimension of the feature vector.
Equation **(14)** helps us in obtaining the eigenvectors and eigenvalues by doing a quantum phase estimation.

With reference to equation **(12)**, we extend $|\tilde{y}_j\rangle$ to obtain the eigenvalues $((\lambda_r)_j)$ and eigenvectors $(|(E_r)_j\rangle)$ of $\hat{F}$ as $|\tilde{y}_j\rangle = \sum_{r=1}^{M_j^{(f,s)}+1} \langle (E_r)_j | \tilde{y}_j \rangle |(E_r)_j\rangle$. Phase estimation generates a state, which stores the respective eigenvalues $|\tilde{y}_j\rangle |0\rangle \to \sum_{l=1}^{M_j^{(f,s)}+1} \frac{\langle (E_r)_j | \tilde{y}_j \rangle}{(\lambda_r)_j} |(E_r)_j\rangle$, where we see an inversion of the eigenvalues and obtain the eigenvalues by performing uncomputing [25] [26] the eigenvalue register and a controlled rotation around it.

Therefore, the $|b_j, \vec{\alpha}_j\rangle$ is obtained by inverting eigenvalues (as discussed in the last paragraph) and expressing $|\tilde{y}_j\rangle$ in the eigenvectors. The overall runtime training complexity for the $j^{th}$ case is $O(logM_j^{(f,s)}N)$, where each of the $M_j^{(f,s)}$ examples are having either $f$ or $s$ as class value.

Here the kernel matrix plays a vital role in the dual formulation of equation **(7)**, and the dot product calculation. In the quantum version of the algorithm, the dot product is calculated quantum mechanically [24]. We calculate a dot product of two training instances as follows: at first, using an ancilla variable we generate two quantum states, $|\psi\rangle$ and $|\varphi\rangle$, then we evaluate the sum of the squared norms of the two input objects. We then compare the two input objects and execute a projective measurement on the ancilla alone.

Let us consider a linear kernel $K_{lin} = x_i^T x_j = \left(\frac{|x_i|^2 + |x_j|^2 - |x_i - x_j|^2}{2}\right)$, for which we estimate the dot product. With the help of QRAM (Quantum Random Access Memory) [27], we construct the quantum state

$$|\psi\rangle = \frac{1}{\sqrt{2}}(|0\rangle|x_i\rangle + |1\rangle|x_j\rangle), \tag{16}$$

and estimate $|\varphi\rangle = \frac{1}{(|x_i|^2+|x_j|^2)}(|x_i||0\rangle - |x_j||1\rangle)$. The quantum state is $\frac{1}{\sqrt{2}}(|0\rangle - |1\rangle) \otimes |0\rangle$, which gets evolved with the Hamiltonian:

$$H = (|x_i||0\rangle\langle 0| + |x_j||1\rangle\langle 1|) \otimes \sigma_x \tag{17}$$

This results in the following state

$$\frac{1}{\sqrt{2}}(\cos(|x_i|t)|0\rangle - \cos(|x_j|t)|1\rangle) \otimes |0\rangle - \frac{i}{\sqrt{2}}(\sin(|x_i|t)|0\rangle - \sin(|x_j|t)|1\rangle) \otimes |1\rangle \tag{18}$$

Measuring the ancilla bit with appropriate t, the complexity of constructing $|\varphi\rangle$ with accuracy $\epsilon$ and $(|x_i|^2 + |x_j|^2)$ is $O(\epsilon^{-1})$. We now perform a swap test on the ancilla alone with $|\psi\rangle$ and $|\varphi\rangle$. Thus, the runtime complexity of calculating a single dot product $x_i^T x_j$ with QRAM is $O(\epsilon^{-1}logN)$. A QRAM uses $n$ qubits to address any quantum superposition of $N$ memory cells. It exponentially reduces the requirements for memory access and needs only $O(log N)$ switches for retrieving the information from the register, where $N = 2^n$ is the feature vector dimension and $n$ is the number of qubits of address register in QRAM.

Suppose, $K_{poly}(x_i, x_j) = \varphi(\vec{x}_i).\varphi(\vec{x}_j) = (x_i, x_j)^d$ is the $d^{th}$ order polynomial kernel. The SVM classification can be performed in the higher dimensional space. In this case, each vector is mapped into a $d$-times tensor product $|\varphi(\vec{x}_i)\rangle \equiv |\vec{x}_i\rangle \otimes ... \otimes |\vec{x}_i\rangle$. The runtime complexity of this quantum polynomial kernel trick is $O(dlogN/\epsilon)$. Apart from the quantum polynomial kernel, we have also discussed the quantum Gaussian kernel in our recent work **[14]**. The runtime complexity of normalized quantum Gaussian kernel **[14]** is $\sim O[\epsilon^{-1}(1 + e)logM] \sim O[\epsilon^{-1}logM]$.

3. CLASSICAL SUPPORT VECTOR CLUSTERING

In **[7]** authors have discussed the support vector clustering (SVC). In the first phase, it defines the one-class support vector machine (SVM) to find the cluster boundaries. One class SVM problem is actually equivalent to finding a minimum region $R$, which encloses most of the input objects.

For our discussion in this paper, at first, following a similar approach as in the SVM classification in the original formulation **[7]**, we formulate the classical support vector clustering using the least squared support vector machine, as compared to the original implementation in **[7]**. Later in Section 4, we show how to formulate a quantum version of it. The implementation with a classical least squared SVM uses a classical kernel trick (with polynomial and/or Gaussian kernel function), whereas the original implementation in **[7]** used the classical Gaussian kernel. The kernel trick transforms the input objects into a higher dimensional space to handle the non-linear dataset. Using a kernel (polynomial or Gaussian) function also helps in forming tight contours in the higher dimensions. The support vectors surround the boundaries of these contours. We represent these contours as clusters. We modified the equation **(11)** to make it feasible for clustering, as follows:

$$(0, \vec{y}^T)^T = F(b, \vec{\alpha}^T)^T \approx f(\vec{x}) = sgn\left(\sum_{\vec{x}_i \in \text{ support vectors}} \alpha_i K(\vec{x}_i, \vec{x}) + b\right) \quad (19)$$

where, $\alpha_i > 0$ corresponds to the support vectors and for rest of the points, $\alpha_i = 0$, and $K$ represents the kernel function. The objective here is to address a one-class SVM for the clustering implementation that finds a minimal region $R$, which encloses the input data objects. By using the input data that has only single class, we construct a one-class SVM classifier from a generalized multiclass SVM. One class SVM deduces the properties of the single class cases and from these properties predict which examples are unlike the given class examples. A positive value outcome of $f(\vec{x})$ implies that $\vec{x}$ falls within the dense subspace $R$, whereas the negative value outcome of the decision function **(19)** implies a sparsely populated region. The objects for which $f(\vec{x}) < 0$ are known as *bounded support vectors (BSV)*, and support vectors are those objects which fall on the contour line. Inside a contour, we have the clustered objects for that specific contour.

In the second phase, the one-class SVM formulation helps in framing the clustering implementation by computing an adjacency matrix $A$ of input objects. In this formulation, $A_{ij} = 1$ if $\vec{x}_i$ and $\vec{x}_j$ are enclosed within the same contour, and 0 otherwise. Equation **(19)** helps in determining whether $\vec{x}_i$ and $\vec{x}_j$ lie within the same contour or not. In this case, Equation **(19)** determines the decision for all the objects on the line that connects $\vec{x}_i$ and $\vec{x}_j$, and if the Equation **(19)** results in all positive values, then that all these objects are within the same contour and bounded by the support vectors $\vec{x}_i$ and $\vec{x}_j$. Each contour is then considered as a separate cluster. Algorithm 1 **[7]** determines the number of clusters and the objects in the clusters:

| **ALGORITHM 1:** Cluster Finding Algorithm |
|---|
| **clusterFinding(A):** |

1. Initialize
   1.1 All the vertices $\vec{x}_i$ in $A$ as not marked,
   1.2 A variable $clusterCount = 1$,
1.3 A one-dimensional dynamic array $storeClusterObjects$
1.4 A multidimensional dynamic array $clusteredObjects$
2. Loop the following for every vertex $\vec{x}_i$ in $A$
   2.1 If $\vec{x}_i$ is not marked, then call $depthFirstSearch(\vec{x}_i)$
   2.2 $clusterCount = clusterCount + 1$
   2.3 Append *storeClusterObject*s to *clusteredObjects*
3. Return *clusterCount, clusteredObjects*

The Algorithm 1, **clusterFinding,** returns the number of clusters, *clusterCount,* and the cluster objects linked with each cluster, *clusteredObjects*.

---

**ALGORITHM 2:** Depth First Search Algorithm

**depthFirstSearch($\vec{x}_i$):**
1. Mark $\vec{x}_i$
2. $storeClusterObjects = \vec{x}_i$
3. Search an adjacency of $\vec{x}_i$, say $\vec{x}_j$, that has not yet been visited using a classical search algorithm
4. Loop the following for every adjacency $\vec{x}_j$ of $\vec{x}_i$.
   4.1 If $\vec{x}_j$ is not marked, then call $depthFirstSearch(\vec{x}_j)$
5. Return *storeClusterObjects*

The Algorithm 2, **depthFirstSearch,** is the depth-first-search (DFS) algorithm. With Gaussian kernel formulation, $K = e^{-\sigma \|\vec{x}_i - \vec{x}_j\|^2}$, let us suppose that $\sigma$ is the scale parameter of the Gaussian kernel and $\gamma$ is the soft margin constant (Please refer *Section 2.1* for more details on soft margin constant). These two parameters govern the contours in the cluster data space. The parameter $\gamma$ controls the number of outliers in the data space, and variations in the value of $\sigma$ may vary the number of clusters in the data space.

4. QUANTUM SUPPORT VECTOR CLUSTERING

The quantum support vector clustering implementation has two phases, similar to the case in the classical counterpart and described in the previous section. In the first phase, we formulate the cluster boundaries with a one-class quantum least square support vector machine and, subsequently, in the second phase, we identify the number of clusters and the objects within the clusters.

*4.1 Cluster Boundaries*

Data preparation and pre-processing in a quantum setup is a complex task. We address the task of representation of classical data into quantum form by using a quantum random access memory (QRAM) **[27]**. Quantum random access memory allows us to perform memory access in coherent quantum superposition access, and thus the data can be accessed in a quantum parallel way **[27] [21]**. With the similar context of classical random access memory, the QRAM is composed of the input register (address register) and the output register, but in qubits instead of bits form, and the memory array can be in quantum or classical form based on the specific use cases. In our quantum setting, all the data inputs are in quantum superposition. The address register *AR* in QRAM contains a superposition of addresses $\sum_l \psi_l |l\rangle_{AR}$, and by correlating with the address register, the QRAM returns the data register *DR,* which contains a superposition of output data:

$$\sum_l \psi_l |l\rangle_{AR} \xrightarrow{QRAM} \sum_l \psi_l |l\rangle_{AR} |D_l\rangle_{DR}, \qquad (20)$$

where the $l^{th}$ memory cell contains $D_l$.

The quantum support vector machine **[13] [1]** is a quantum classification algorithm that works based on the postulates of quantum mechanics. In our construction, we first need a one-class quantum SVM. We articulate a quantum SVM formulation into a one-class quantum SVM formulation, which is a straightforward process with a simple trick. We just use the data for which all the input objects have the same labels. Suppose, $|\vec{x}_i\rangle$ and $|\vec{x}_j\rangle$ are two input objects. We determine whether the $|\vec{x}_i\rangle$ and $|\vec{x}_j\rangle$ lie within the same contour, by classifying all the objects using the following quantum SVM decision function **[13] [14]**,

$$|\vec{y}_{one\_class\_case}\rangle = \hat{F}_{one\_class\_case}(|b_{one\_class\_case}, \vec{\alpha}_{one\_class\_case}\rangle), \qquad (21)$$

on the line that connects $|\vec{x}_i\rangle$ and $|\vec{x}_j\rangle$. The $+1$ classification for all the objects guarantees that $|\vec{x}_i\rangle$ and $|\vec{x}_j\rangle$ are in the same contour. Here, the equation **(21)** is the simplified one-class representation of the general quantum SVM **(12)** & **(13)**. Using the quantum equation **(21)**, we can similarly perform the operation we do with equation **(19)** (i.e, the classical case). In this quantum space, we obtained the SVM parameters $\vec{\alpha}$ and $b$ using quantum mechanical postulates, where, $\vec{\alpha} > 0$ corresponds to the support vectors and for rest of the points, $\vec{\alpha} = 0$.

## *4.2 Cluster Identification*

After formulating the quantum one-class support vector machine, our next task is to assign the cluster boundaries. The single class quantum SVM can be easily extended to a clustering scheme by computing an adjacency matrix *A* for the given cluster data, where:

$$A_{ij} = \begin{cases} 1; \text{ if } |\vec{x}_i\rangle \text{ and } |\vec{x}_j\rangle \text{ are enclosed within the same contour.} \\ 0; \text{ otherwise.} \end{cases} \qquad (22)$$

By classifying all the points in the line that connects $|\vec{x}_i\rangle$ and $|\vec{x}_j\rangle$, we determine whether $|\vec{x}_i\rangle$ and $|\vec{x}_j\rangle$ lie within the same contour. All the objects within a specific contour form a separate cluster. Technically, the number of contours means the number of clusters. In the graph induced by the matrix $A_{ij}$, one can detect the connected components. The number of clusters is determined by the number of graphs induced by the connected components in $A_{ij}$. Let us discuss finding the number of connected components in $A_{ij}$. For this purpose, we have performed a depth-first-search (DFS) in a quantum way **[28]**. Here, $A_{ij}$ is an undirected graph matrix. The following Algorithm 3 determines the number of clusters and associated objects in the clusters:

**ALGORITHM 3:** Quantum Cluster Finding Algorithm

**quantumClusterFinding(A):**
1. Initialize
    1.1 All the vertices $|\vec{x}_i\rangle$ in $A$ as not marked,
    1.2 A variable $clusterCount = 1$,
    1.3 A one-dimensional dynamic array *storeClusterObjects*
    1.4 A multidimensional dynamic array *clusteredObjects*
2. Loop the following for every vertex $|\vec{x}_i\rangle$ in $A$
    2.1 If $|\vec{x}_i\rangle$ is not marked, then call *quantumDepthFirstSearch*($|\vec{x}_i\rangle$))
    2.2 $clusterCount = clusterCount + 1$
    2.3 Append *storeClusterObject*s to *clusteredObjects*
3. Return *clusterCount, clusteredObjects*

**ALGORITHM 4:** Quantum Depth First Search Algorithm

**quantumDepthFirstSearch($x_i$):**
1. Mark $|\vec{x}_i\rangle$
2. $storeClusterObjects = |\vec{x}_i\rangle$
3. Search an adjacency of $|\vec{x}_i\rangle$, say $|\vec{x}_j\rangle$, that has not yet been visited using a quantum search algorithm (Grover's quantum search in this case)
4. Loop the following for every adjacency $|\vec{x}_j\rangle$ of $|\vec{x}_i\rangle$.
	4.1 If $|\vec{x}_j\rangle$ is not marked, then call *quantumDepthFirstSearch($|\vec{x}_j\rangle$)*
5. Return *storeClusterObjects*

Algorithm 3, **quantumClusterFinding,** returns the number of clusters, *clusterCount,* and the cluster objects associated with each cluster, *clusteredObjects*. The Algorithm 4, **quantumDepthFirstSearch,** is the quantum version of depth-first-search (QDFS) algorithm.

With the quantum version of a Gaussian kernel, the contours in the cluster data space are governed by two parameters, $\sigma$, and $\varsigma$, where $\sigma$ is the scale parameter of quantum Gaussian kernel and $\varsigma$ is the soft margin constant. The shape/number of the boundaries in the data space varies with the changes in $\sigma$. Increasing the value of $\sigma$ may result in increasing the number of clusters in the data space. The parameter $\varsigma$ controls the number of outliers in the data space.

5. COMPUTATIONAL COMPLEXITY AND ERROR ANALYSIS

At first, we analyze the complexity of the classical implementation with the least square SVM. We discuss the complexity of the whole approach for finding the *clustering boundaries* and the *clustering identification* phases, respectively. We start the discussion with the *clustering boundary* phase, in the case of the least square SVM, quadratic programming is circumvented, and the parameters are evaluated from the solution of a system of linear equations. In classical settings **[29]**, the algorithm converges after approximately $O(M^2)$ kernel evaluations. The complexity of the complete algorithm is $O(M^2N)$, which is a polynomial-time complexity, and where the number of support vectors is $O(1)$. In the *clustering identification* phase, the complexity of finding the number of clusters in the adjacency matrix $A$ with *depth-first-search* is $O(M^2)$.

In the quantum SVM paradigm, the performance gain in the dimensional factor $N$ is due to the fast quantum evaluation of inner products. We achieve the performance advantage in the number of training examples $M$ by re-expressing the SVM formulation as an approximate least square formulation, which allows us to employ the matrix inversion algorithm **[13]** and using a technique for the non-sparse matrices exponentiation. Assuming, $\epsilon_K$ is the smallest eigenvalue measured and $\epsilon$ is the accuracy, the training stage error dependence is $(poly(\epsilon_K^{-1}, \epsilon^{-1}))$ **[11] [12]**. When a low-rank approximation is appropriate, the quantum SVM runs on the complete training set in logarithmic runtime.

In the quantum setting for support vector clustering, we initially discuss the complexity of finding *cluster boundaries.* For $N$ dimensional $M$ cluster data points, the least square quantum SVM classification training with a quantum linear kernel takes $O(logNM)$ runtime complexity. When implemented with the quantum Gaussian kernel, the runtime complexity of the kernel implementation is approximated $O[\epsilon^{-1}(1+e)logN]$, where $\epsilon$ is the maximal error in the context. And, with the quantum $d$-level polynomial kernel, the runtime complexity of the quantum support vector machine is $O(d\epsilon^{-1}logN)$.

The runtime complexity of the *clustering identification* phase considers two runtime contributions. During the quantum search **[29]** in quantum DFS, the search will fail once **[30]**, which includes a runtime complexity of $O(\sqrt{M^3 log_2 M})$. Now, we consider the runtime complexity of the successful search **[31]** of the element, which is also $O(\sqrt{M^3 log_2 M})$. Therefore, the runtime complexity of the *clustering identification* phase is

($\sqrt{M^3 log_2 M}$) with quantum implementation. In this quantum setting, $\epsilon_g^{-1}$ is polynomial in the number of vertices of the adjacency matrix, where we suppose that the anticipated probability of failure is $\epsilon_g$ **[28]**.

When we transfer the classical data onto the quantum form, the classical data with $N$ dimensional complex form can be mapped onto a quantum state over $log_2 N$ qubits, the runtime computational complexity of this mapping is $O\ (log_2 N)$. Thus, QRAM takes only $O\ (log_2 N)$ steps to query the memory for reconstructing a state. This computational complexity adds extra $O\ (log_2 N)$ factor to the overall complexity, when we orchestrate the support vector clustering algorithm in the quantum system. $O\ (log_2 N)$ is small to affect the overall complexity in quantum paradigm when compared to the overall complexity in the classical paradigm.

We see that in both phases, i.e., for *finding clustering boundaries* and *cluster identification,* we achieved near exponential and quadratic performance gains, respectively. Therefore, the overall runtime analysis concludes that the proposed quantum version implementation of the support vector clustering is significantly faster than the classical implementation.

6.  SIMULATION BASED PRACTICAL STUDIES

*6.1 Practical Exploration*

Our proposed quantum support vector clustering approach shows significant speedup gain theoretically as compared to the classical counterpart, and the use of quantum version of SVM and Grover's quantum search play the most vital role in this achievement. **Fig.1** exhibits the runtime scaling between the classical and quantum SVM against the number of training examples (for simplicity assuming the dimension of the feature vector approximately equal to the number of training examples). **Fig.2** shows the runtime scaling between the classical and quantum search against the number of inputs.

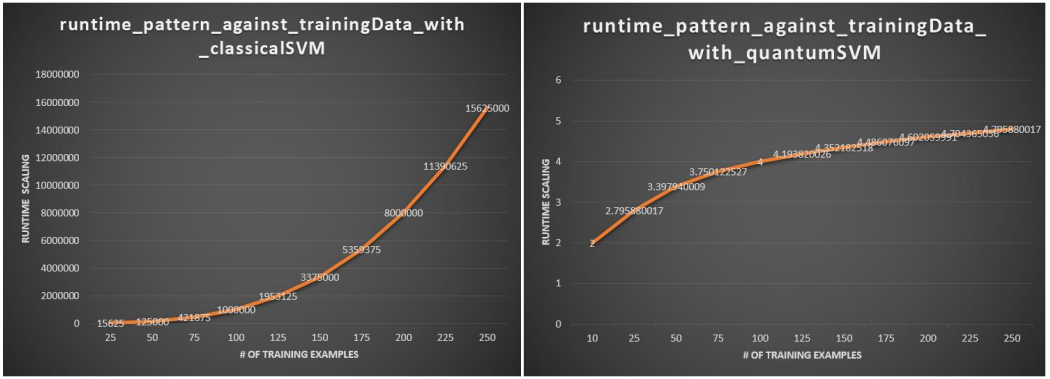

*Fig.1: Runtime comparison between classical SVM ($O(M^2 N) \approx O(M^3)$) vs quantum SVM ($O(log\ MN)$) against the number of training examples*

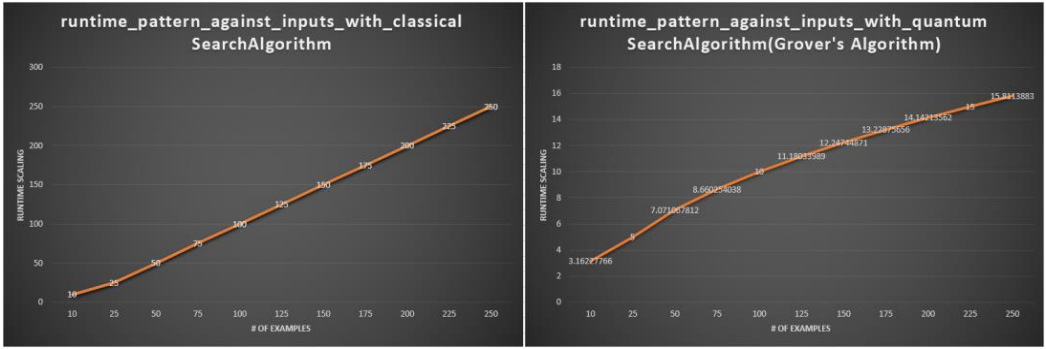

*Fig.2: Runtime comparison between classical search $(O(M))$ vs Grover's search $\left(O(\sqrt{M})\right)$ against the number of inputs*

We are here dealing with big data. The number of qubits required for implementing quantum SVM is directly proportional to the dimension of the feature vector. Therefore, for $N$ dimensional feature vectors, we need the $N$ qubits quantum system. Although, in technical implementation, in special consideration, we can reduce the size of the feature dimension by using a dimension reducing technique - for example "Principal component analysis" (PCA), which is also an interesting topic to discuss for any advantage gain in overall accuracy against the less number of features. But here, we are talking about big data only. For example, an image of size $150 \times 150$ pixels requires $150 \times 150 = 22,500$ qubits system to process 22,500 features. Although we can reduce the number of feature vectors before feeding it into the SVM by using convolutional techniques, still we require very high volume-qubit quantum systems. At the present time, no quantum computer supports such a large number of qubits. In image processing, the image sizes may even vary from $150 \times 150 \times 3$ to $400 \times 400 \times 3$ pixels (when taking into account the third color channel too). Apart from this, QRAM (Quantum random access memory) helps in mapping the classical $N$ dimensional feature vector over $log_2 N$ qubits **[27]** too.

For implementing the Grover's search algorithm to search a marked item in the list of $M$ items, we need $log_2 M$ qubits system. So, for implementing the second part of our proposed quantum SVM clustering algorithm, we need at least a $log_2 M$ qubits quantum system, where $M$ is the number of training samples

For the demonstration purpose, to investigate our quantum SVM clustering approach in a simulated quantum computer, we examined the quantum SVM algorithm in the IBMQ **[32]** quantum simulator (with qasm_simulator). We use the QISKIT library **[33]** to implement the simulation environment. We used the *Breast cancer* dataset **[34]**. As we used a two qubits system, we have transformed the 30-dimensional space of the feature vectors to only two-dimensional space of feature vectors using PCA (principal component analysis). Here, we have to keep the dimension of the feature vectors equal to the number of qubits. The experimental results are presented in the below table Table 1. In a similar way, we can execute with more than two qubits quantum system.

Table.1

| Models | Simulation environment setup time (seconds) | Quantum circuit building time (seconds) | Training time (seconds) | Prediction time (seconds) | Test accuracy |
|---|---|---|---|---|---|
| classical SVM | 0 | 0 | 0.000993 | 00.001487 | 0.99 |
| qasm_simulator | 00.007441 | 30 | 0.006000 | 0.0000600 | 0.90 |

The above Table 1 illustrates that we are still far away from experiencing the true power of the quantum computer. The simulated environment is slow as compared to the classical one for training with the same data sets. Although we observed better performance during the prediction with quantum simulated environment. We tested with only 20 training examples and with 2 qubits setup for the analysis purpose. Further there is restriction on the maximum number of qubits in IBMQ, which is limited to 4, 5 and ~53 qubit systems, which are still very small scale quantum computers and hardly useful for state-of-the-art machine learning tasks). The experimental results are hoped to be near to the theoretical analysis when we will have more sophisticated quantum computer with much higher number of qubits. Similarly, Grover's algorithm [35] can be implemented with three qubits with IBMQ, but will not have any practical use of it as the number of available qubits is too small.

## 6.2 Implementation Inadequacy

The QISKIT interface has a limitation of ~75 circuits support. This puts a restriction on the multiple iteration support for Grover's search [36]. Although, the IBMQ group is working on high volume qubits systems (a recent one with ~53 qubits). But, as the number of qubits increases, it attracts more execution time and gate errors due to gate coupling, which may affect our algorithm's performance in terms of overall accuracy. It will be interesting as a future investigation to measure the algorithm's performance with every addition of the new qubits in IBM's system. With the quantum simulator, it is possible to configure a coupling map equal to that of the IBMQ quantum computer. However, in case of the physical qubits, each qubit may interact differently with other qubits and quantum gates, due to the limitation of the hardware and coupling map layout in the present quantum systems. Therefore, to test our algorithm with big data, we need a quantum computer with a very high volume of qubits (at least equal to the number of feature vectors) and with a very good error correction mechanism. Quantum decoherence is another factor, which needs to be addressed while designing a large quantum computer to support the proposed quantum algorithm which deals with big data. The effects of the decoherence increase as the number of qubits increases.

## 7. CONCLUSIONS

Clustering is used in an assortment of applications such as document clustering, market segmentation, and image segmentation, etc. The the idea is to get some evocative intuition of the structure of the data we're dealing with and group them based on thier similarities. The support vector clustering (SVC) is one of the most popular clustering method which is based on support vector machine. In the investigation in this paper, we have discussed on the theoretical grounds that the SVC implementation in the quantum paradigm exhibits better than quadratic speed up gain in overall performance as compared to the classical implementation. We have analyzed the quantum implementation of the support vector clustering method with both quantum Gaussian kernels as well as with quantum polynomial kernels, and have concluded that both implementations have shown substantial performance improvements in the overall runtime complexity as compared to the classical implementation. The Gaussian/polynomial kernels help in developing better contours in higher dimensions both in the classical implementation [37] as well as in the quantum version. Training of the one-class SVM is also exponentially faster in its quantum version. During the clustering identification phase, we have used a quantum version of the depth-first search, which shows quadratic speed up gain as compared to the classical implementation of DFS. DFS is used to identify the number of clusters and the clustered objects, with the help of the adjacency matrix. In the proposed quantum version of support vector clustering approach, we have demonstrated significant quantum advantages on performance gains at multiple stages, i.e., the one class SVM formulation, the kernel formulation, and during the depth-first search with Grover's search algorithm. We have also discussed the implementation possibilities at the present time with the IBMQ quantum computer and concluded that the implementation of our proposed quantum algorithm with big data requires a quantum computer with very high volume of qubits.